# Multicuts and Perturb & MAP for Probabilistic Graph Clustering

Jörg Hendrik Kappes · Paul Swoboda · Bogdan Savchynskyy · Tamir Hazan · Christoph Schnörr

**Abstract** We present a probabilistic graphical model formulation for the graph clustering problem. This enables to locally represent uncertainty of image partitions by approximate marginal distributions in a mathematically substantiated way, and to rectify local data term cues so as to close contours and to obtain valid partitions.

We exploit recent progress on globally optimal MAP inference by integer programming and on perturbation-based approximations of the log-partition function, in order to sample clusterings and to estimate marginal distributions of node-pairs both more accurately and more efficiently than state-of-the-art methods. Our approach works for any graphically represented problem instance. This is demonstrated for image segmentation and social network cluster analysis. Our mathematical ansatz should be relevant also for other combinatorial problems.

## 1 Introduction

Clustering, image partitioning and related NP-hard decision problems abound in the fields image analysis, computer vision, machine learning and data mining, and much research has been done on alleviating the combinatorial difficulty of such inference problems using various forms of relaxations. A recent assessment of the state-of-the-art using discrete graphical models has been provided by [20]. A subset of the specific problem instances considered there (Potts-like functional minimisation) are closely related to continuous formulations investigated, e.g., by [11,27].

From the viewpoint of statistics and Bayesian inference, such *Maximum-A-Posteriori (MAP)* point estimates have been always criticised as falling short of the scope of probabilistic inference, that is to provide – along with the MAP estimate – "error bars" that enable to assess sensitivities and uncertainties for further data analysis. Approaches to this more general objective are less uniquely classified than for the MAP problem. For example, a variety of approaches have been suggested from the viewpoint of clustering (see more comments and references below) which, on the other hand, differ from the variational marginalisation problem in connection with discrete graphical models [42]. From the computational viewpoint, these more general problems are even more involved than the corresponding MAP(-like) combinatorial inference problems.

In this paper, we consider graph partitioning in terms of the minimal cost multicut problem [12], also known as correlation clustering in other fields [6], which includes modularity clustering [10], the image partitioning problem [4] and other graph partition problems [32] as special case. Our work is based on

(i) recent progress [34,16] on the probabilistic analysis of perturbed MAP problems applied to our setting in order to establish mathematically the connection to basic variational approximations of inference problems [42],
(ii) recent progress on *exact* [22,23] and *approximative* [8] solvers of the minimum cost multicut problem, which is required in connection with (i).

Figure 1 provides a first illustration of our approach when applied for image partitioning. Instead of only calculating the most likely partition, our approach additionally provides alternative probable partitions and returns quantitative measures of certainty of the boundary parts.

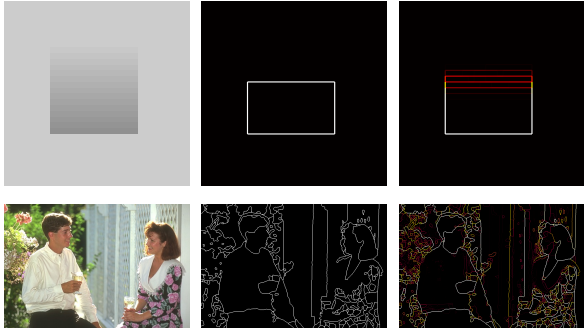

**Fig. 1** Two examples demonstrating our approach. **Left column:** images subject to unsupervised partitioning. **Center column:** globally optimal partitions. **Right column:** probabilistic inference provided along with the partition. The color order: white → yellow → red → black, together with decreasing brightness, indicate uncertainty, cf. Fig. 3. We point out that *all local* information provided by our approach is intrinsically *non-locally* inferred and relates to partitions, that is to *closed* contours.

Although probabilistic image partitioning has been motivating our work, the resulting approach is more widely applicable. This is demonstrated in the experimental section by analyzing a problem instance from the field of machine learning in terms of network data defined on a general graph.

### 1.1 Related Work

The susceptibility of clustering to noise is well known. This concerns, in particular, clustering approaches to image partitioning that typically employ spectral relaxation [39, 19, 28]. Measures proposed in the literature [29, 32] to quantitatively assess confidence in terms of stability, employ data perturbations and various forms of cluster averaging. While this is intuitively plausible, a theoretically more convincing substantiation seems to be lacking, however.

In [18], a deterministic annealing approach to the unsupervised graph partitioning problem (called pairwise clustering) was proposed by adding an entropy term weighted by an artificial temperature parameter. Unlike the simpler continuation method of Blake and Zisserman [9], this way of smoothing the combinatorial partitioning problem resembles the variational transition from marginalisation to MAP estimation, by applying the log-exponential function to the latter objective [42]. As in [9], however, the primary objective of [18] is to compute a single "good" local optimum by solving a sequence of increasingly non-convex problems parametrised by an artificial temperature parameter, rather than sampling various "ground states" (close to zero-temperature solutions) in order to assess stability, and to explicitly compute alternatives to the single MAP solution. The latter has been achieved in [33] using a non-parametric Bayesian framework. Due to the complexity of model evaluation, however, authors have to resort to MCMC sampling.

Concerning continuous problem formulations, a remarkable approach to assess "error bars" of variational segmentations has been suggested by [36]. Here, the starting point is the "smoothed" version of the Mumford-Shah functional in terms of the relaxation of Ambrosio and Tortorelli [2] that is known to $\Gamma$-converge to the Mumford-Shah functional in the limit of corresponding parameter values. Authors of [36] apply a particular perturbation ("polynomial chaos") that enables to locally infer confidence of the segmentation result. Although being similar in scope to our approach, this approach is quite different. An obvious drawback results from the fact that minima of the Ambrosio-Tortorelli functional do not enforce partitions, i.e. may involve contours that are not closed.

Finally, we mention recent work [37] that addresses the same problem using – again – a quite different approach: "stochastic" in [37] just refers to the relaxation of binary indicator vectors to the probability simplex, and this relaxation is solved by a *local* minimisation method.

All the previous approaches require to select or optimize over an unknown number of clusters. This introduces a bias into the model, as we will show later. Our approach, on the other hand, works on an exponential family over edge-variables. It models clusterings in terms of multicuts, which inherently includes selection of the number of clusters. Samples from an approximation of this distribution are obtained as MAP-solutions of randomly perturbed partition problems.

### 1.2 Basic Notation

For the set of natural numbers from 1 to $k$ we use the shorthand $[k]$ and denote the cardinality of a set $A$ by $|A|$. For a set $A$ we denote by $[V]^k$ the set of $k$-element subsets of $A$ and for the sets $A_1$ and $A_2$ by $[A_1, A_2]$ the set $\{\{a_1, a_2\} | a_1 \in A_1, a_2 \in A_2\}$. By $2^A$ we denote the power set of $A$, which is defined as the set of all subsets of $A$. The inner product of vectors is denoted by $\langle \cdot, \cdot \rangle$, and the indicator function $\mathbb{I}(expression)$ is 1 if the expression is true and 0 otherwise. We use the classical notation for a undirected graph $G = (V, E)$ where $V$ is a set of nodes and $E \subset [V]^2$ is a set of edges. The degree of a node $v$ is given by $\deg(v) := |\{u \in V : \{u, v\} \in E\}|$.

1.3 Organization

The remaining paper is organized as follows. We will give the formal definition of partitioning, the minimal cost multicut problem and its polyhedral representation in Sec. 2. This is followed by the definition of probabilistic distributions over partitions and methods that estimate marginals and generate samples in Sec. 3. A detailed numerical evaluation on synthetic and real world data is given in Sec. 4.

## 2 Graph Partitioning and Multicuts

The minimal cost multicut problem, also known as *correlation clustering*, is defined in terms of partitions of an undirected weighted graph

$$G = (V, E, w), \quad V = [n], \quad E \subseteq [V]^2, \tag{1a}$$
$$w \colon E \to \mathbb{R}, \qquad e \mapsto w_e := w(e) \tag{1b}$$

with a signed edge-weight function $w$. A positive weight $w_e > 0$, $e \in E$ indicates that it is beneficial to put the two nodes into the same cluster, whereas a negative weight indicates that it is beneficial to separate them. We formally define below valid partitions and interchangeably call them clusterings.

**Definition 1 (partition, clustering)** A set of subsets $\{S_1, \ldots, S_k\}$, called *shores*, *components* or *clusters*, is a *(valid) partition* of a graph $G = (V, E, w)$ iff (a) $S_i \subseteq V$, $i \in [k]$, (b) $S_i \neq \emptyset$, $i \in [k]$, (c) the induced subgraphs $G_i := (S_i, [S_i]^2 \cap E)$ are connected, (d) $\bigcup_{i \in [k]} S_i = V$, (e) $S_i \cap S_j = \emptyset$, $i, j \in [k]$, $i \neq j$. The set of all valid partitions of $G$ is denoted by $\mathcal{S}(G)$.

The number $|\mathcal{S}(G)|$ of all possible partitions is upper-bounded by the Bell number [1] that grows very quickly with $|V|$.

**Definition 2 (minimal cost multicut problem)** The *correlation clustering* or *minimal cost multicut problem* is to find a partition $S^*$ that minimizes the cost of intra cluster edges as defined by the weight function $w$.

$$S^* \in \arg\min_{S \in \mathcal{S}(G)} \sum_{ij \in E} w_{ij} \sum_{k=1}^{|S|} \mathbb{I}(i \in S_k \wedge j \notin S_k) \tag{2}$$

The minimal cost multicut problem can be formulated as a node labeling problem given by the problem of minimizing a Potts model

$$x^* \in \arg\min_{x \in V^{|V|}} \sum_{ij \in E} w_{ij} \mathbb{I}(x_i \neq x_j). \tag{3}$$

Since any node can form its own cluster, $|V|$ labels are needed to represent all possible assignments in terms of variables $x_i$, $i \in V$. From an optimizer $x^*$ of (3) we can get an optimizer $S^*$ of (2) by calculating the connected components on $G' = (V, \{ij \in E \colon x_i = x_j\})$.

A major drawback of this formulation is the huge space needed to represent the assignments. Furthermore, due to the lack of an external field (unary terms), any permutation of an optimal assignment results in another optimal labeling. As a consequence of this symmetry, the standard relaxation in terms of the so-called local polytope [42] becomes too weak and can not handle the necessary non-local constraints, cf. Sec.3.1.

In order to overcome these problems, we adopt an alternative representation of partitions based on the set of *inter* cluster edges as suggested in [12]. We call the edge set

$$\delta(S_1, \ldots, S_k) := \bigcup_{i \neq j,\, i,j \in [k]} [S_i, S_j] \cap E \tag{4}$$

a *multicut* associated with the partition $S = \{S_1, \ldots, S_k\}$. To obtain a polyhedral representation of multicuts, we define for each subset $E' \subseteq E$ an *indicator vectors* $\chi(E') \in \{0,1\}^{|E|}$ by

$$\chi_e(E') := \begin{cases} 1, & \text{if } e \in E', \\ 0, & \text{if } e \in E \setminus E'. \end{cases}$$

The *multicut polytope* $\mathcal{MC}(G)$ then is given by the convex hull

$$\mathcal{MC}(G) := \operatorname{conv}\{\chi(\delta(S)) \colon S \in \mathcal{S}(G)\}. \tag{5}$$

The vertices of this polytope are the indicator functions of valid partitions and denoted by

$$\mathcal{Y}(G) := \{\chi(\delta(S)) \colon S \in \mathcal{S}(G)\}. \tag{6}$$

Based on this representation, the minimal cost multicut problem amounts to find a partition $S \in \mathcal{S}(G)$ that minimizes the sum of the weights of edges cut by the partition

$$\arg\min_{S \in \mathcal{S}(G)} \sum_{e \in E} w_e \cdot \chi_e(\delta(S))$$
$$\equiv \arg\min_{y \in \mathcal{MC}(G)} \sum_{e \in E} w_e \cdot y_e. \tag{7}$$

This problem is known to be NP-hard [6] and moreover APX-hard [13]. Although problem (7) is a linear program, the representation of the multicut polytope $\mathcal{MC}(G)$ by half-spaces is of exponential size and moreover, unless $P = NP$, no efficient separation procedure for the complete multicut polytope exist [14].

However, one can develop efficient separation procedures for an outer relaxation of the multicut polytope which involves all facet-defining cycle inequalities. Together with integrality constraints, this guarantees globally optimal solutions of problem (7) and is still applicable for real world problems [21,20]. For the tractability of huge models and to provide better anytime performance, several greedy move-making methods [24,5,8,7] has been suggested. These methods are able to find nearly optimal solutions for larger models much faster.

## 3 Probabilistic Graph Partitioning

Additionally to the most likely partitioning, one might also be interested in the probability of partitions and the probability that a certain edge is part of a cut. For this we define for a given graph $G$ a probability distribution over all partitions in terms of an exponential family

$$p(y|\theta) := \exp\left(\langle\theta, y\rangle - A(\theta)\right) \qquad \theta := \frac{-w}{T} \qquad (8)$$

$$A(\theta) := \ln\left(\sum_{y \in \mathcal{Y}(G)} \exp\left(\langle\theta, y\rangle\right)\right), \qquad (9)$$

where $w \in \mathbb{R}^{|E|}$ is the vector due to the edge weights (1b) and $\mathcal{Y}(G)$ is given by (6). $T \geq 0$ is called *temperature parameter*. This exponential family differs from usual discrete exponential families in the log partition function $A(\theta)$. Instead of calculating the sum over $\{0,1\}^{|E|}$, as it is usual for discrete graphical models, we implicitly take into account the topological constraints encoded by the multicut polytope by restricting the feasible set to $\mathcal{Y}(G)$.

While the MAP-inference problem considers to find $y^*$ that maximizes $p(y|\theta)$, it is also useful to do probabilistic inference and calculate the probability that an edge $e \in E$ is cut and thus contributes to separating different clusters. This probability is given by the marginal distributions

$$p(y_e|\theta) := \mathbb{E}_\theta(y_e) = \sum_{y' \in \mathcal{Y}(G), y'_e = y_e} p(y|\theta). \qquad (10)$$

Since $p(y|\theta)$ is an exponential family we know [42] that $A(\theta)$ is a convex function of $\theta$ and

$$\mathbb{E}_\theta(y_e) = \frac{\partial A(\theta)}{\partial \theta_e}, \qquad (11)$$

The (Legendre-Fenchel) conjugate function of $A(\theta)$ is given by

$$A^*(\mu) = \sup_{\theta \in \mathbb{R}^{|E|}} <\theta, \mu> -A(\theta) \qquad (12)$$

and takes values in $\mathbb{R} \cup \infty$. we call $\mu \in \mathbb{R}^{|E|}$ *dual variables*. Theorem 3.4 in [42] gives an alternative interpretation of $A^*(\mu)$ as the negative entropy of the distribution $p_{\theta(\mu)}$ with $\mathbb{E}_{\theta(\mu)}[y] = \mu$, where $\theta = \theta(\mu)$ is defined by the right-hand side of (12),

$$A^*(\mu) = \begin{cases} -H(p_{\theta(\mu)}) & \text{if } \mu \in \mathcal{MC}^\circ(G) \\ \lim_{n \to \infty} -H(p_{\theta(\mu^n)}) & \text{if } \mu \in \mathcal{MC}(G) \setminus \mathcal{MC}^\circ(G) \\ \infty & \text{if } \mu \notin \mathcal{MC}(G) \end{cases}$$

Here $\mathcal{MC}^\circ(G)$ is the relative interior of $\mathcal{MC}(G)$, and $\mu^n$ a sequence in $\mathcal{MC}^\circ(G)$ that converges to $\mu$. The log partition function has the variational representation

$$A(\theta) = \sup_{\mu \in \mathcal{MC}(G)} \{\langle\theta, \mu\rangle - A^*(\mu)\} \qquad (13)$$

and the supremum in Eq. (13) is attained uniquely at the vector $\mu \in \mathcal{MC}(G)^1$ specified by the moment-matching condition $\mu = \mathbb{E}_\theta(y)$.

Thus, calculating the marginals amounts to solving the convex problem (13). This is at least as hard as solving the MAP-inference problem, which requires to solve a linear program, since both the convex set $\mathcal{MC}(G)$ and the entropy $-A^*(\mu)$ are not tractable in general.

We will discuss next how to approximate this problem by variational approaches and perturb & MAP.

### 3.1 Variational Approach

Since solving problem (13) is NP-hard, we need tractable approximations of $\mathcal{MC}(G)$ and of the negative entropy $A^*(\mu)$. Common variational approximations for unconstrained graphical models consider outer relaxations on the feasible set and tractable approximations on the entropy in terms of pseudo-marginals, which are then solved by message passing algorithms.

Although formulation (3) has been used before in the literature (e.g., [38]), we have to cope with several particular issues. Firstly, contrary to [38] our model includes also negative couplings, which renders the problem more challenging. Secondly, the label-space can become very large when using a formulation in the node-domain, as done in (3)[2]. Finally, the most crucial problem is that there is not a one-to-one correspondence between a clustering and a node-labeling of the nodes. For example, if we have $k$ labels, then k node-labelings exists to represent the clustering into one class. On the other hand, any clustering into 2 clusters can be modeled by $k \cdot (k-1)$ node-labelings. As a consequence we would take some partitions more often into account than others during marginalisation. This ambiguity can be partially reduced by assigning the label 1 to the first node w.l.o.g. as suggested in [38], however ambiguities between remaining labels are still present if $k > 2$. Even exact probabilistic inference by the junction tree algorithm (JTA-n) then does not provide correct marginals, cf. Figs. 4 and 5. When using approximative inference, in this case loopy belief propagation (LBP), the error increases.

---
[1] For $\theta \in \mathbb{R}^N$ we have $\mu \in \mathcal{MC}^\circ(G)$. Boundary points of $\mathcal{MC}(G)$ are only reached when at least one entry of $\theta$ is infinity.
[2] To overcome this problem we will restrict the number of possible labels and exclude thereby some partitions. If the number of used labels is greater or equal than the chromatic number of the graph all partitions are representable.

To avoid these ambiguities, we switch to graphical models in the edge domain, as defined by Eq. (8). As an outer relaxation of the multicut polytope we consider the cycle-multicut polytope $\mathcal{MC}_C(G)$. It is defined as the intersection of half-spaces given by the cordless cycles of $G$ denoted by $C(G) \subset 2^E$:

$$\mathcal{MC}_C(G) = \{[0,1]^{|E|} \mid \forall e \in C \in C(G) : \sum_{e' \in C \setminus \{e\}} y_{e'} \geq y_e\}.$$

As shown by Chorpa [12] we have $\mathcal{Y}(G) = \mathcal{MC}_C(G) \cap \{0,1\}^{|E|}$. So $\mathcal{MC}_C(G)$ is a efficiently separable outer relaxation of $\mathcal{Y}(G)$ and consequently an outer relaxation for $\mathcal{MC}(G)$, too. Accordingly, we define an alternative exponential family

$$p(y|\bar{\theta}) := \exp\left(\langle \bar{\theta}, \phi(y)\rangle - A(\bar{\theta})\right) \quad (14)$$

$$A(\bar{\theta}) := \ln\left(\sum_{y \in \{0,1\}^{|E|}} \exp\left(\langle \bar{\theta}, \phi(y)\rangle\right)\right) \quad (15)$$

with the exponential parameter $\bar{\theta}$ and the sufficient statistic $\phi(y)$ both in $\mathbb{R}^{|E|+\sum_{C \in C(G)} 2^{|C|}}$ and defined by

$$\bar{\theta}_e = \frac{-w_e}{T} \quad (16)$$

$$\bar{\theta}_{C,y_C} = \begin{cases} 0 & , \text{if } \forall e \in C : \sum_{e' \in C \setminus \{e\}} y_{e'} \geq y_e \\ -\infty & , \text{otherwise} \end{cases} \quad (17)$$

$$\phi(y)_e = y_e \quad (18)$$

$$\phi(y)_{C,y'_c} = \mathbb{I}(y'_C \equiv y_C) \quad (19)$$

Contrary to the formulation in Eq. (8), the constraints are encoded in the exponential parameter, which is that why marked by an additional bar.

It is worth to mention that the dimension of the exponential parameter vector grows exponentially with the length of the longest non-chordal cycle. If $G$ is chordal[3], then $C(G)$ includes only a polynomial number of cycles of length three, which can be easily enumerated. If $G$ is not chordal, we have to deal factors of high order, i.e. the objective includes terms that depend on more than two variables, which is practically

---

[3] A graph is called chordal, if each cycle of length larger 3 have a chord, which is an edge that is not part of the cycle but connects two vertices of the cycle.

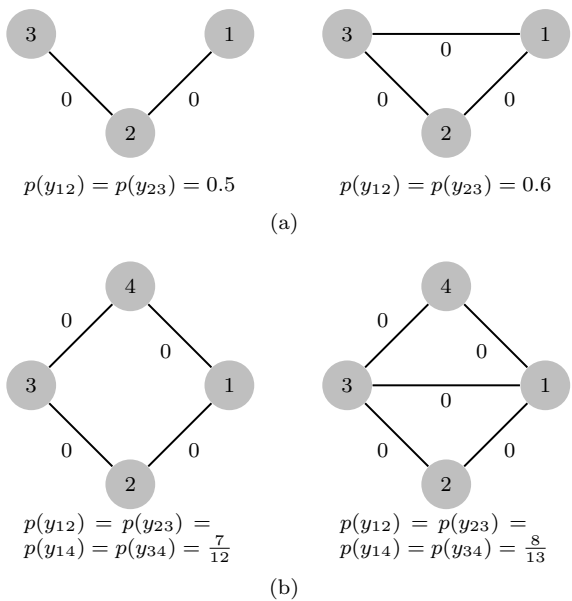

$p(y_{12}) = p(y_{23}) = 0.5$    $p(y_{12}) = p(y_{23}) = 0.6$

(a)

$p(y_{12}) = p(y_{23}) =$    $p(y_{12}) = p(y_{23}) =$
$p(y_{14}) = p(y_{34}) = \frac{7}{12}$    $p(y_{14}) = p(y_{34}) = \frac{8}{13}$

(b)

**Fig. 2** Adding zero weighted edges to a graph can change the likelihood of partitions and marginal distribution for cut edges. In example (a), the additional edge destroys the independence of random variables for edge-cuts by enforcing topological consistency and forbids to cut only one of the 3 edges. This causes a bias towards a cut. Similarly, in example (b), the triangulation of the graph causes a bias towards cutting edges. The cut $(1,1,1,1)$ in the left graph has two valid corresponding cuts $((1,1,1,1,0),(1,1,1,1,1))$ in the right graph.

not tractable. To overcome this problem, we add zero-weighted edges in order to make the graph chordal. While the corresponding factor graph has then order three, an additional bias is introduced into the model, cf. Fig. 2.

The advantage of this ansatz is that eq. (14)-(19) define a discrete graphical model, for which several methods exists to estimate the marginals. This includes the junction tree algorithm (JTA) [26] for exact inference on graphs with small tree-width, as well as loopy belief propagation (LBP) [25], tree re-weighted belief propagation (TRBP) [41], and generalized belief propagation (GBP) [43] for approximative inference by message passing. In addition to the above-mentioned bias for non-chordal graphs, the approximative methods suffer from the combinatorial nature of the higher-order terms (17) that can not be dealt with by local messages, cf. Sec. 4.1.3.

### 3.2 Perturbation & MAP

Recently, Hazan and Jaakkola [16] showed the connection between extreme value statistics and the partition function, based on the pioneering work of Gumbel [15].

In particular they provided a framework for approximating and bounding the partition function as well as creating samples by using MAP-inference with randomly perturbed models.

Analytic expressions for the statistics of a random MAP perturbation can be derived for general discrete sets, whenever independent and identically distributed random perturbations are applied to every assignment.

**Theorem 1 ([15])** *Given a discrete distribution $p(z) = \exp(\theta(z) - A(\theta))$ with $z \in \mathcal{Z}$ and $\theta : \mathcal{Z} \to \mathbb{R} \cup \{-\infty\}$, let $\Gamma$ be a vector of i.i.d. random variables $\Gamma_z$ indexed by $z \in \mathcal{Z}$, each following the Gumbel distribution whose cumulative distribution function is $F(t) = \exp(-\exp(-(t+c)))$ (here c is the Euler-Mascheroni constant). Then*

$$\Pr[\hat{z} = \arg\max_{z \in \mathcal{Z}} \{\theta(z) + \Gamma_z\}] = \exp(\theta(\hat{z}) - A(\theta)),$$

$$\mathbb{E}[\max_{z \in \mathcal{Z}} \{\theta(z) + \Gamma_z\}] = A(\theta).$$

Theorem 1 offers a principled way based on solving the MAP problem for randomly perturbed model parameters, to compute the log partition function (13) in view of computing the marginals (11) as our objective. For larger problems, however, the number of states is too large and thus Thm. 1 not directly applicable. Hazan and Jaakkola [16] developed computationally feasible approximations and bounds of the partition function based on *low*-dimensional random MAP perturbations.

**Theorem 2 ([16])** *Given a discrete distribution $p(z) = \exp(\theta(z) - A(\theta))$ with $z \in \mathcal{Z} = [L]^n$, $n = |V|$ and $\theta : \mathcal{Z} \to \mathbb{R} \cup \{-\infty\}$. Let $\Gamma'$ be a collection of i.i.d. random variables $\{\Gamma'_{i;z_i}\}$ indexed by $i \in V = [n]$ and $z_i \in Z_i = [L]$, $i \in V$, each following the Gumbel distribution whose cumulative distribution function is $F(t) = \exp(-\exp(-(t+c)))$ (here c is the Euler-Mascheroni constant). Then*

$$A(\theta) = \mathbb{E}_{\Gamma'_{1;z_1}}\left[\max_{z_1 \in \mathcal{Z}_1} \cdots \mathbb{E}_{\Gamma'_{N;z_n}}\left[\max_{z_n \in \mathcal{Z}_n} \theta(z) + \sum_{i \in [n]} \Gamma'_{i;z_i}\right]\right].$$

Note that the random vector $\Gamma'$ includes only $nL$ random variables. Appying Jensen's inequality, we arrive at a computationally feasible upper bound of the log partition function [16],

$$A(\theta) \leq \mathbb{E}_{\Gamma'}\left[\max_{z \in Z} \theta(z) + \sum_{i \in [n]} \Gamma'_{i;z_i}\right] =: \tilde{A}(\theta). \quad (20)$$

The samples that are generated by the modes of the perturbed models follow the distribution

$$\tilde{p}(\hat{z}|\theta) = \Pr[\hat{z} \in \arg\max_{z \in Z} \theta(z) + \sum_{i \in [n]} \Gamma'_{i;z_i}]. \quad (21)$$

It has been reported [35,16,17] that this distribution $\tilde{p}$ is a good approximation for $p$, but contrary to the fully perturbed model it is not known so far how to explicitly represent $\tilde{p}$ as member of the exponential family of distributions. This precludes a analytic evaluation of this approximation.

### 3.3 Perturb & MAP for Graph Partitioning

In the following we will review the possible sampling schemes with Perturb & MAP, based on Sec. 3.2, for the graph partition problem described in Sec. 2. Accordingly, we denote again the random variables by $y$ instead of $z$.

#### 3.3.1 Unbiased Sampling

Based on Theorem 1 we can define an *exact* sampler for the distributions of partitions for *small* graphs. For each possible partition we have to add a random variable. The number of possible partitions of $N$ elements is given by the Bell number. In our setting it can be smaller, since the partition has also to respect the graph topology and single clusters have to be connected. In prior work [23] we showed how to include such higher-order terms into the MAP-inference. Sampling a partition from $p(y|\theta)$ reduces to solving

$$\arg\max_{y \in Y(G)} \langle \theta, y \rangle + \sum_{y \in Y(G)} \gamma_y \quad \gamma_y \sim Gumbel(c,1), \quad (22)$$

where $\gamma_y$ is a sample from a Gumbel distribution with location $c$ and scale 1 denoted by $Gumbel(c,1)$. Note that contrary to any MCMC-algorithm, which generates true samples only in the limit (after convergence of a Markov chain), samples that follow the original distribution can be generated by solving the combinatorial problem (22) in finite time.

#### 3.3.2 Fast Biased Sampling

When the low-dimensional perturbation as in Thm. 2 is used, samples are generated by

$$\arg\max_{y \in Y(G)} \langle \theta, y \rangle + \sum_{e \in E} (\gamma_{e;1} \cdot y_e + \gamma_{e;0} \cdot (1 - y_e)) \quad (23)$$

$$= \arg\max_{y \in Y(G)} \langle \theta, y \rangle + \sum_{e \in E} (\gamma_{e;1} - \gamma_{e;0}) \cdot y_e \quad (24)$$

$$\gamma_{e;y_e} \sim Gumbel(c,1)$$

Note that the difference of two independent random variables that follow a Gumbel distribution follows a logistic distribution [40].

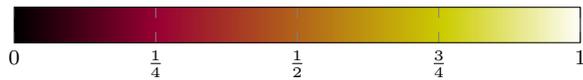

**Fig. 3** The colormap used in this paper for visualizing the probability that an edge ($E$) is cut $Pr[y_e = 1]$

In order to generate of a sample, a minimal cost multicut problem with perturbed weights has to be solved. This can be done either exactly by integer linear programming [23] or approximatively [24,8]. The latter option introduces an additional bias and does no longer guarantee that eq. (21) holds. Empirically we observe however that the additional error is small for our problems, as the approximate solutions are near optimal and negligible compared to the gap introduced by the Jensen's inequality in eq. (21).

The samples follow the distribution $\tilde{p}(\hat{y}|\theta)$ for which the log-partition function is given by $\tilde{A}(\theta)$, cf. Eqns. (21), (21). Consequently, in view of (11) the marginals $\tilde{\mu}$ of $\tilde{p}(y|\theta)$ approximate the marginals $\mu$ of the original distribution (8) by

$$\mu \approx \tilde{\mu} = \nabla_\theta \tilde{A}(\theta) \qquad (25a)$$
$$= \mathbb{E}_{\Gamma'}\left[\arg\max_{y \in Y(G)}\left\{\langle\theta, y\rangle + \sum_{e \in E}\Gamma'_{e;y_e}\right\}\right] \qquad (25b)$$
$$\approx \frac{1}{M}\sum_{k=1}^{M}\arg\max_{y \in Y(G)}\left\{\langle\theta, y\rangle + \sum_{e \in E}\gamma'^{(n)}_{e;y_e}\right\}, \qquad (25c)$$
with $\gamma'^{(n)}_{e;y_e} \sim \Gamma'_{e;y_e}$.

## 4 Experiments

We compare the following methods indicated by acronyms in bold font.

Bruteforce (**BF**) enumerates all valid partitions. This is only practicable for small graphs up to ten nodes, but provides globally optimal unbiased marginals.

The local estimates (**LOCAL**) ignore topological constraints and compute marginals by

$$Pr[y_e = 1] = \frac{\exp(-w_e)}{\exp(-w_e) + \exp(0)}.$$

This method can be used as a baseline to measure the improvements by the use of topology.

The junction tree algorithm (**JTA**) can be applied in the node domain (**JTA-n**), cf. Sec. 3.1. We use 4 labels which is sufficient to represent each partition of a planar graph and a reasonable trade-of between expressiveness and computational cost for general graphs. Alternatively, in the edge-domain JTA can be applied on the triangulated graph with additional triple-constraints (**JTA-e**), cf. Sec. 3.1. The latter method computes the optimal marginals if the original graph is chordal. Since the complexity of JTA is exponential in the tree-width it does not scale well. We used the implementation within libDAI [31].

Alternatively to JTA we tested loopy belief propagation (**LBP**) in the node- (**LBP-n**) and edge-domain (**LBP-e**), again with the publicly available implementation of libDAI [31]. While LBP has no convergence guarantees it shows good convergence behavior if damping is used. In the node domain the probability of a cut is calculated as the sum of terms with different labels in the pairwise marginals. We also try generalized belief propagation [43] and tree reweighted belief propagation [41], which are more theoretically substantiated. They often failed however due to numerical problems caused by the constraint-terms.

For the proposed Perturb & MAP approach we consider global Perturb & MAP (**G-P&M**) for tiny models, as described in Eq. (22), as well as low-dimensional Perturb & MAP, as described in Eq. (23), together with exact inference by [23] (**L-P&M**) or approximative inference by CGC [8] (**L-P&M-CGC**) and Kernighan Lin [24] (**L-P&M-KL**), for solving the perturbed problems. For all P&M methods we used OpenGM [3] to solve the subproblems. For estimating the marginals, we used 1000 samples for synthetic models and 100 samples for real-world models.

All experiments were done on machines equipped with an Intel Core i7-2600K (3.40GHz) CPU, 8 MB cache and 16 GB RAM, with a single thread.

### 4.1 Synthetic Models

For evaluations in a controlled setting we generated fully connected graphs and grid graphs with edge-weights that are independently normal distributed with mean zero and variance 1. As long as the graphs are small, exact marginals can be calculated by enumerating all edge-labelings in $\{0,1\}^{|E|}$ or all multicuts $\mathcal{Y}(G)$ which can be precomputed for a given graph $G$. The corresponding results for the fully connected and grid graphs are shown in Fig 4 and 5, respectively.

For the fully connected graphs JTA-e is exact and JTA-n has small error in the marginals. Since both methods would not scale to larger graphs we resorted to LBP in the node and edge domain. In both cases the estimated marginals became worse as the graph size grows. L-P&M gave slightly worse results on the average, but produced no outliers. Using the KL-heuristic (L-P&M-KL) instead of exact inference did not effect the quality much. For grid models even exact inference

JTA-e and JTA-n returned results worse than L-P&M for graphs larger than $3 \times 3$, caused by the inherent bias of the underlying approximations. Again, the use of approximative sub-solvers in L-P&M-CGC did not deteriorate results compared to exact inference for these problem sizes. By and large the error increases less for L-P&M than for JTA and LBP.

### 4.1.1 Effect of Graph-topology

As illustrated in Fig. 2 the topology of a graph has an impact on the probability that an edge is cut and a graph with zero weights have not necessary a probability 0.5 that an edge is cut. Adding edges to a graph introduce a bias, which we can observe in Fig. 5. The junction tree algorithm (JTA-e) calculates the optimal marginals on the triangulated graph. However, these marginals can be far from the exact marginals of the non-triangulated graph. For larger graphs this error can be larger than the error caused by low-dimensional perturbation, as it is the case for $3 \times 3$ and $4 \times 4$ grids.

The experiments furthermore show, that the choice of the topology already has an influence on the likelihood that an edge is cut. In real world applications that should be considered when e.g. selecting super-pixels for image segmentation.

### 4.1.2 Effect of Temperature

The temperature parameter $T$ controls the standard deviation of the distribution and the impact of the perturbation. When the temperature drops to 0 all the mass of the distribution concentrates on the optimal solutions and samples are drawn uniformly from these "ground" states. When the temperature grows to $\infty$ the mass covers the entire feasible set and samples are drawn uniformly from all valid segmentations. Fig. 6 illustrates this for a discrete distribution. For low temperatures it becomes extremely unlikely that the perturbation is strong enough to "lift" a non-optimal state above the optimal one. For large temperatures however the difference in the original model are negligible compared to the perturbation variance.

From a theoretical point of view the temperature parameterizes a non-linear transformation of the marginals. From a practical viewpoint, it controls the variance of the samples. In other words, if we could sample infinitely many clusterings we could recalculate the marginals for any density. With a finite number of samples we can adjust the resolution for more or less likely edge-cuts. For example less likely cuts become visible for high temperatures (at the cost of more noise) in

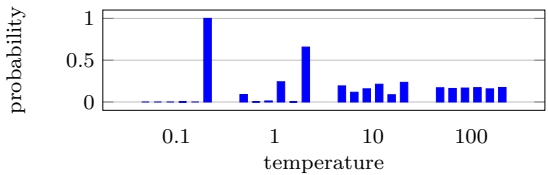

**Fig. 6** Discrete distributions $p(i) \propto \exp(-\theta_i/T)$ for $\theta = (2, 7, 4, 1, 10, 0)$ and different temperatures. For $T \to \infty$ the distributions becomes uniform, for $T \to 0$ all its mass concentrates on a single mode.

Fig. 9 while those are not visible for low temperatures when using the same number of samples.

As a rule of thumb we found that the temperature should be selected such that on the average to 10% of the variables the optimal label is not assigned. This generates samples that are diverse enough, but not too diverse.

### 4.1.3 Evaluation with Ground-truth

An exact evaluation of the marginal distributions is only possible for very small graphs for which we can compute the log-partition function by enumerating all partitions. As fully connected graphs are chordal, the junction tree algorithm can then be used. However, its complexity is not less than enumerating all states. Message passing methods are exact for graphs with 3 nodes. For larger graphs the underlying outer relaxation and entropy approximations are no longer tight and get worse with larger graphs. While marginals produced with the Perturbed & MAP technique have a slightly bigger mean error for full connected graphs, Perturb & MAP does not create outliers as the message passing methods. For grid graphs larger than $3 \times 3$ Perturbed & MAP produces better results in terms of mean and maximal error. The use of approximative multicut solvers, namely KL and CGC, for Perturb & MAP leads to similar results as when using exact solvers, but for the synthetic datasets the difference is negligible.

### 4.1.4 Number of Samples

Perturb & MAP generates samples from a distribution. An natural question is how many samples are required for a robust estimation of the marginals. The answer to this question obviously depends on the distribution we are sampling from.

Fig. 7 shows the behavior for a fully connected graph with five nodes (K5). While the error for exact Perturb & MAP (G-P&M) vanishes, the low-dimensional Perturb & MAP (L-P&M) has a systematic error since it does not sample from the original distribution. For this

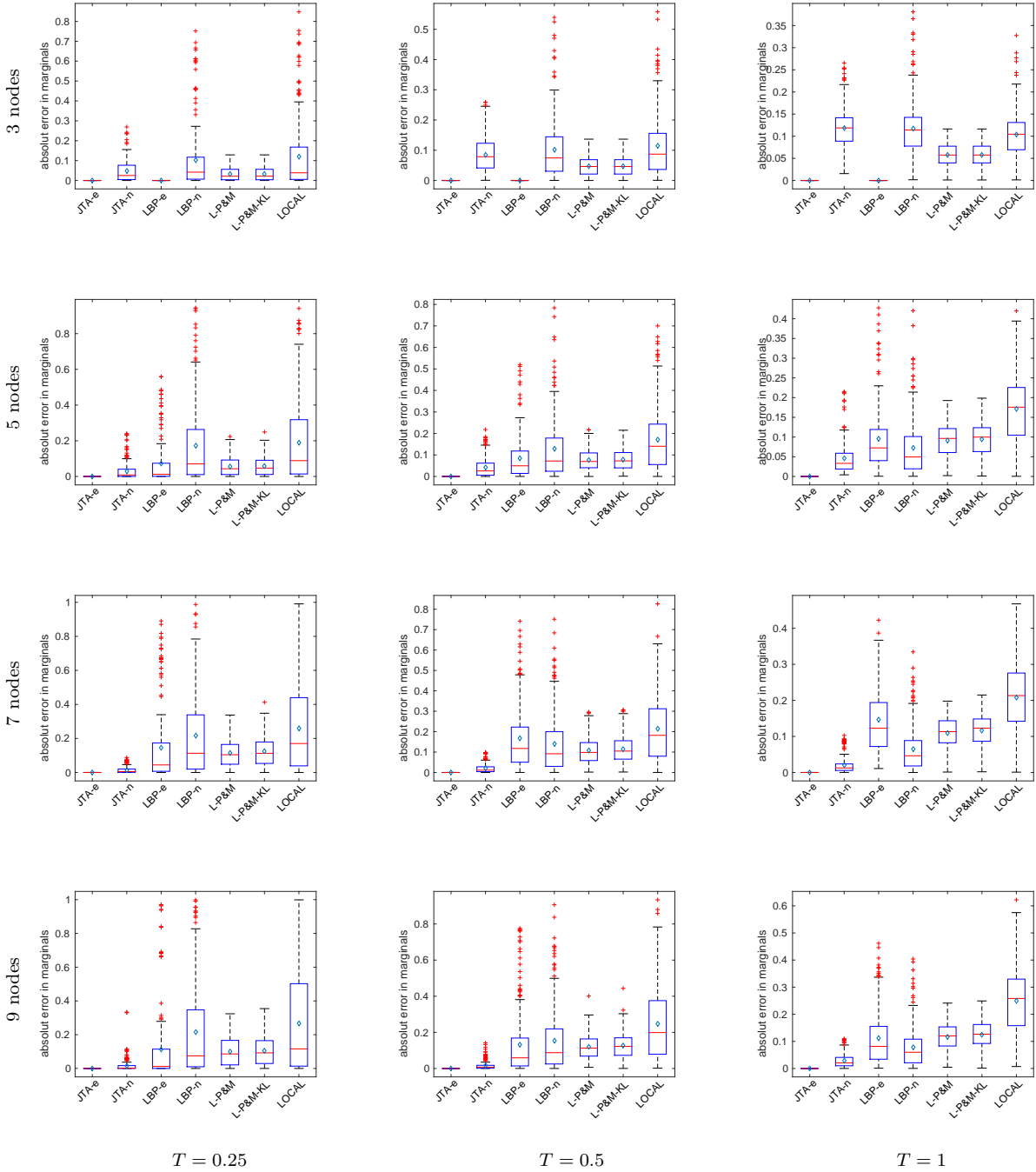

$T = 0.25$          $T = 0.5$          $T = 1$

**Fig. 4** Evaluation of the absolute error of computed marginals for fully connected graphs with 3,5,7, and 9 nodes. For larger models we are no longer able to calculate the exact marginals. The edge weights were sampled from the normal distribution with mean 0 and variance 1. Boxplots show the mean (diamond) and median (redline). The edges of the boxes are the 25th and 75th percentiles. Outliers, with respect to an 1.5 IQR, are marked by red crosses. The black lines mark the largest and lowest non-outliers.
While for 3 nodes LBP-e is exact, L-P&M already have a systematic bias caused by low-order perturbation. For graphs with more than 3 nodes, LBP-e starts suffering from its underlying relaxation. LBP-n produces many bad estimates. L-P&M and L-P&M-KL provide marginals with better or equal accuracy as LBP and no marginals with large errors like its competitors, except JTA-n and JTA-e, which do not scale, however.

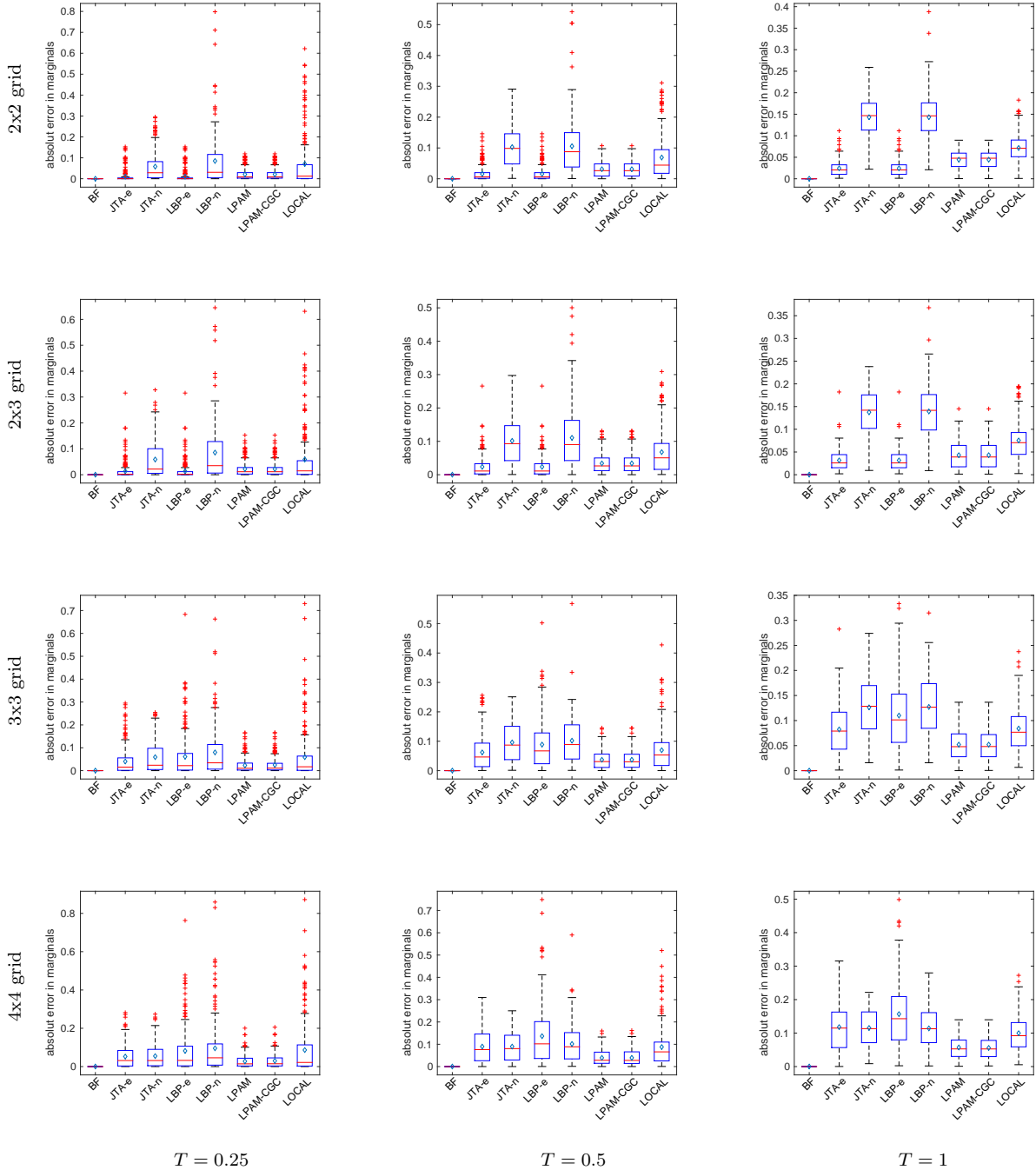

**Fig. 5** Evaluation of the absolute error of computed marginals for grid graphs with 4, 6, 9 and 16 nodes. For larger models we are no longer able to calculate the exact marginals. The edge weights are sampled from a normal distribution with mean 0 and variance 1. Boxplots show the mean (diamond) and median (redline). The edges of the boxes are the 25th and 75th percentiles. Outliers, with respect to an 1.5 IQR, are marked by red crosses. The black lines mark the largest and lowest non-outliers. JTA-e and LBP-e are not exact and include some bias caused by the additional edges included for triangulation. While for small models this error is moderate, it increases for larger grids. L-P&M and L-P&M-CGC have a systematic bias caused by low-order perturbation, which does not grow too much when the grid size increases.

instance after a few thousand samples the mean error in the marginals improves slowly. LBP gives for this small model a better mean error than L-P&M, which is consistent with Fig. 4. Contrary to JTA and LBP L-P&M and G-P&M can be used to calculate higher order marginals, i.e. the likelihood that a set of edges is cut simultaneously.

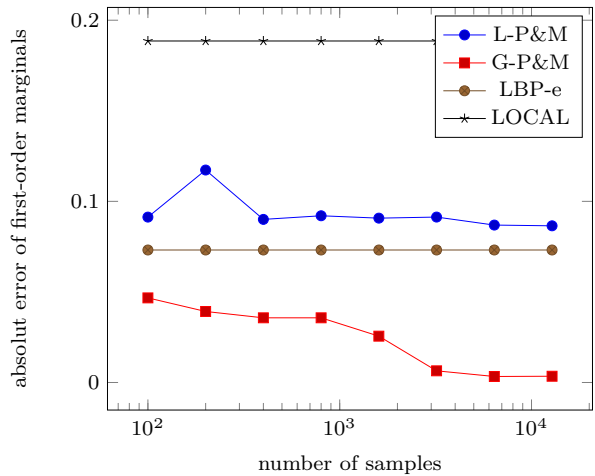

**Fig. 7** Mean absolute error of marginals for the complete graph K5 with normally distributed weights and increasing number of samples. While the error of G-P&M goes to zero, L-P&M and LBP-e have a systematic error that does not vanish. But the estimated marginals are better than the locally computed pseudo marginals.

4.2 Real World Examples

*4.2.1 Image Segmentation*

For image segmentation we use the models of Andres et al. [4], which is based on the Berkeley segmentation dataset [30] and includes 100 instances. Based on a super-pixel segmentation by watershed segmentation, edge-weights of the adjacency graph are calculated as the negative logarithm of the local cut likelihoods given random forest trained on local image features.

$$w_e = -\log\left(\frac{Pr[y_e = 1|\text{feat}_e(I)]}{Pr[y_e = 0|\text{feat}_e(I)]}\right) \quad (26)$$

A coarser clustering than the super-pixels is obtained by calculating the minimal weighted multicut on the weighted adjacency graph. The size of adjacency graph for these instances is too large for applying LBP. So we can only evaluate the results obtained by low-dimensional Perturb & MAP. Due to the lack of probabilistic ground truth and reliable measurements for cut probabilities, which do not form a segmentation and exclude the BSD-measurements [30], we resort to a visual evaluation.

Fig. 8 shows seven exemplary images. The choice of using of exact inference or CGC for the generation of P&M-samples does not effect the marginals much, but CGC is faster and scales much better than the exact method. While the optimal multicut visualizes the most probable partition, the marginals also show uncertainty of the boundary and alternative solutions. The marginals give a better response for windows of the buildings, face of the statue and leafs of the plants compared to the optimal multicut. With increasing temperatures it is more likely to sample boundaries with higher edge weights and reproduce fine structures without strong support of the local data-terms. Visually the models tend to oversegment the image, which does lead to best results in terms of standard measurements for the optimal multicut.

Exemplary samples for different temperatures are shown in Fig. 9. The first images shows the marginal estimated by L-P&M. The remaining 21 images are samples generated by L-P&M. For low temperatures the samples are similar to each other and to the optimal multicut. For higher temperatures the variability of the samples and their cost of the cuts increase. Furthermore we observe, that in regions with small positive edge weights local clusters pop up randomly. However, the main boundary always has a high marginal probability.

*4.2.2 Social Nets*

Another field of application is the clustering of social networks, where individuals are represented by nodes and edges encode if there is a connection between two individuals. Given such a social network $G = (V, E)$, one would like to find a clustering $\mathcal{S}$ of $G$ with maximal modularity $q_G(\mathcal{S})$, that is

$$q_G(\mathcal{S}) := \sum_{S \in \mathcal{S}} \left[\frac{|E \cap [S]^2|}{|E|} - \left(\frac{\sum_{c \in S} deg(v)}{2 \cdot |E|}\right)^2\right] \quad (27)$$

As shown by Brandes et al. [10] on can compute the clustering with maximal modularity in terms of the minimal multicut on a graph $G' = (V, V \times V)$ with weights

$$w_{uv} = \frac{1}{2|E|} \cdot \left(\mathbb{I}(uv \in E) - \frac{\deg(u)\deg(v)}{2 \cdot |E|}\right) \quad (28)$$

As networks we consider a standard dataset which encodes if members of a karate club know each other, and a new dataset which encodes the program committee members of the SSVM 2015 conference as nodes.

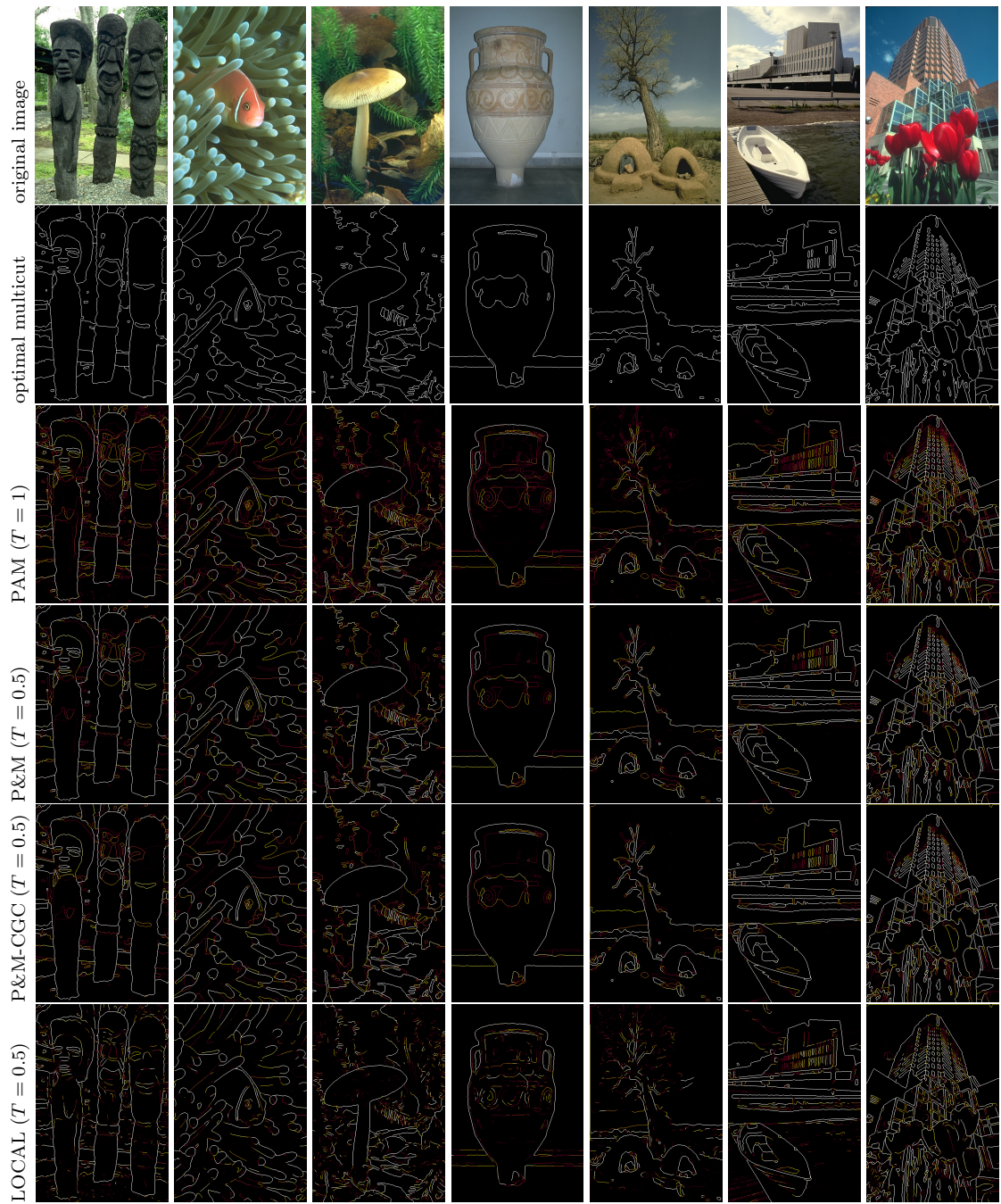

**Fig. 8** Merely computing the optimal partition ignores important information about alternative high-quality partitions that is provided by the marginal distributions. For example some eyes of the statues are not well partitioned and the marginals show that in this region the partition is uncertain. The same holds for the leafs of the tree and the painting on the vase. For the building marginals are able to partition smaller windows. Overall, with larger temperatures the less likely boundaries become more visible. The use of approximative inference (CGC) does not lead to significant changes of the marginals, but speed up computation. Compared to pure local marginals, noisy detections are removed and boundary gaps are closed. Contrary to the local pseudo marginals, the P&M-marginals form a convex combination of valid partitions.

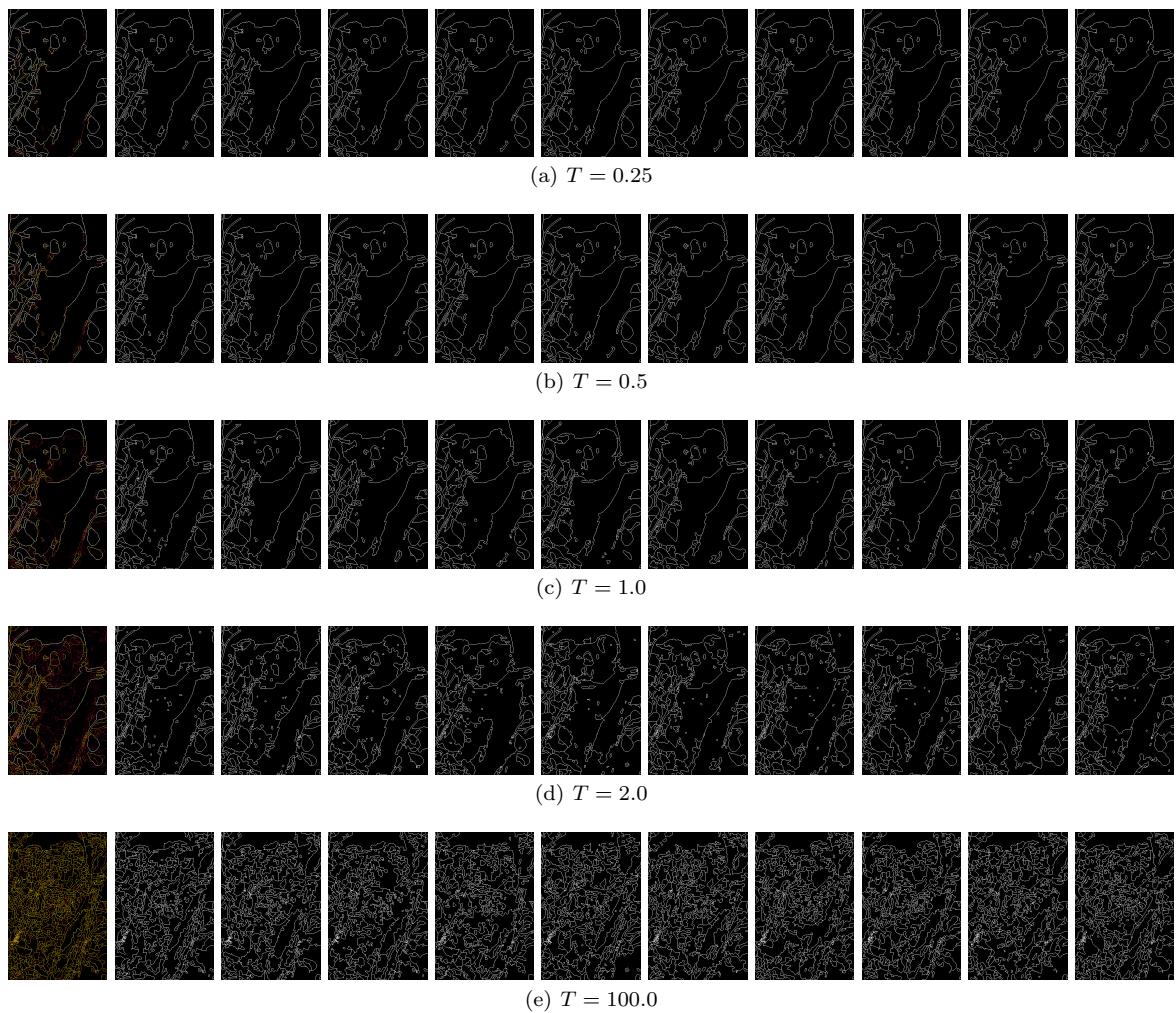

**Fig. 9** Illustration of the marginals and 10 samples for the coala image with temperatures 0.25, 0.5, 1, 2, and 100. For low temperatures the samples are similar to the MAP-solution. For higher temperatures partitions with higher energy get sampled more likely. If the temperature goes to zero only the mode will be sampled. If temperature goes to infinity samples are drawn uniformly from the set of all valid partitions if we would use global perturbation, for low-dimensional perturbation this does not hold exactly.

Edges indicate that the two program committee members had a joint publication listed in the DBLP[4].

Fig. 10 and 11 show the result for the karate instance. The MAP-solution is a partitioning of the graph into 4 clusters as shown in Fig. 11. The local likelihoods (LOCAL) visualized in Fig. 10 represent only local information which does not directly render the most likely clustering or the marginal distribution of edges to be cut. LBP does not work well in the node- and edge-domain as in can not scope with the non-local constraints. L-P&M provides reasonable marginal estimates with exact and approximative sub-solvers. Fig. 11 shows the probabilities of edges to be a cut edge. L-P&M is able to detect the two karate club members (marked by red circles) that could be moved into another cluster without much decreasing the modularity. With higher temperature, less likely partitions becomes relatively more likely. LBP got stuck in local fix-points, which are not sensitive to the temperature.

Fig. 12 shows the result for the SSVM-2015 program committee graph. Contrary to the karate club instance we know the people corresponding to the nodes[5]. This enables us to evaluate the results not only in terms of modularity. Fig. 12(a) shows the joint publication graph extracted from DBLP. For six members of the program committee no joint publications with another program committee member exists. The remaining graph is connected. The optimal clustering of this graph in terms of modularity is shown in Fig. 12(b). The five main clusters are coherent in that its members either are located in a certain region of Europe and/or are members of a scientific subcommunity (scale space, mathematical imaging, computer vision). When sampling clusterings from the corresponding distribution, with temperature $10000^{-1}$ and $50000^{-1}$ additional cluster between the main clusters show up and the marginal distribution for some edges to be cut that are between the main clusters becomes smaller than 1. Each of these edges looks natural in that everyone who knows the community at large (as do the authors), would agree that the two persons that become connected are close to each other, for one or another reason. Such conclusions quite to the point are surprising in view of the fact that the approach only used the DBLP publication data as input.

---

[4] http://dblp.uni-trier.de/db/

[5] For reasons of anonymity, however, we show anonymized results.

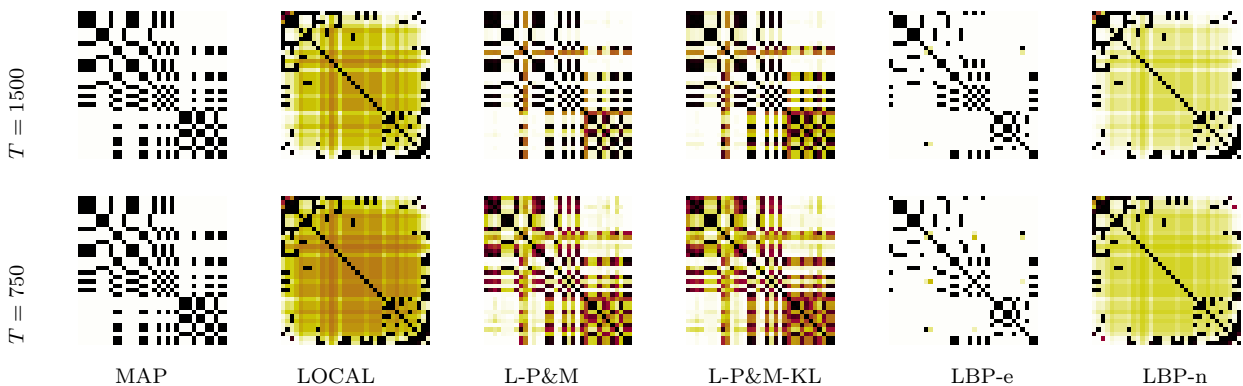

**Fig. 10** The probability that two members belong to the same cluster is visualized by colors range from white to black for increasing probabilities, cf. 3. The MAP-solution results in a hard clustering independent of the temperature. LOCAL provides the local information given by the edge-weights which is the input for the other algorithms. With L-P&M we can identify the two members who can not be assigned to a cluster with high certainty, i.e. they can be assigned to another cluster without decreasing the modularity much. With the lower temperature, partitions with decreasing modularity are sampled more likely. While L-P&M-KL produces similar results compared to L-P&M, LBP-e and LBP-n did not return meaningful results.

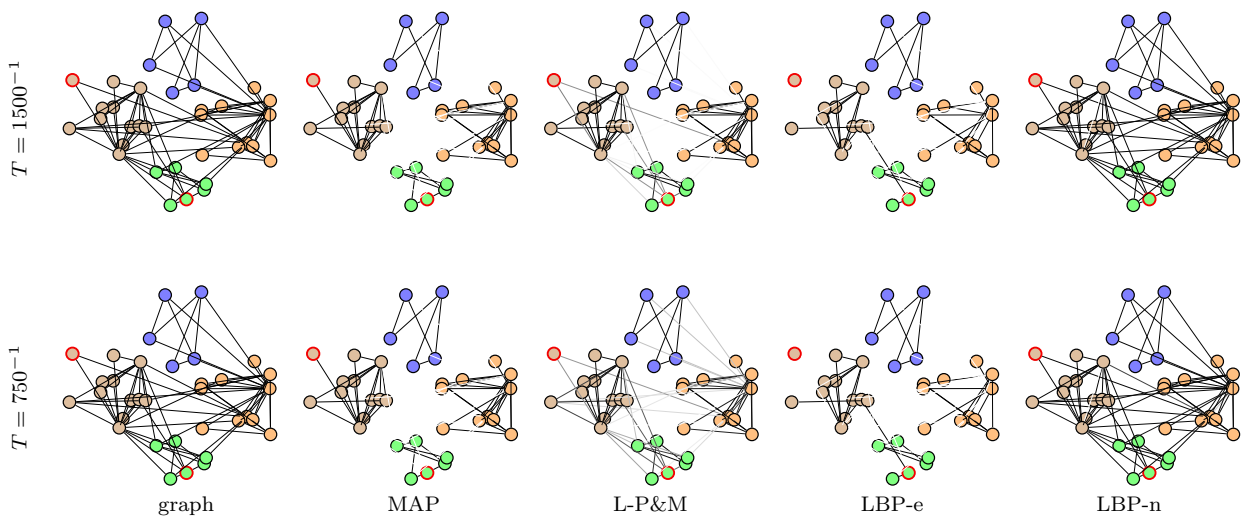

**Fig. 11** Visualization of the clustering of the original adjacency graph. The node colors show the optimal clustering. Nodes with red border are the two most uncertain nodes, that could be moved to another cluster without decreasing the modularity much. Edges colors indicate if an edge is likely to be cut (white) or not (black). Contrary to LBP-e and LBP-n, P&M provides reasonable marginals.

## 5 Conclusions

We presented a novel framework for calculating uncertainty measurements for graph partition problems and a method to generate samples from probability distributions over partitions by solving perturbed minimal multicut problems. Contrary to standard variational approaches, our method can incorporate non-local constraints, does not get stuck in local optima and return reasonable marginal distributions for intra cluster edges. Furthermore our ansatz inherits from the underling multicut framework, that the number of clusters needs not be specified, i.e. the number of clusters is determined as part of the sampling process and different samples may result in different number of clusters.

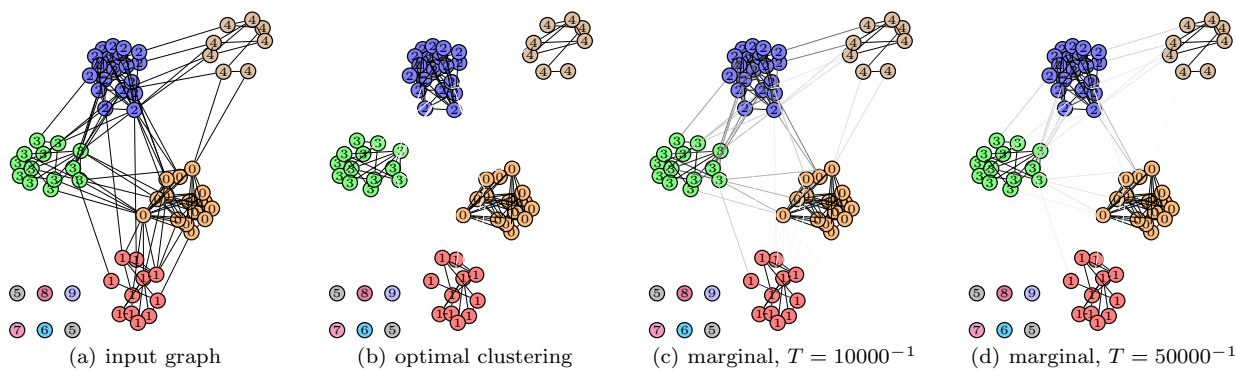

(a) input graph  (b) optimal clustering  (c) marginal, $T = 10000^{-1}$  (d) marginal, $T = 50000^{-1}$

**Fig. 12** The nodes of the graph correspond to the program committee members of the SSVM 2015 conference. Edges indicates that two members have a joint publication in DBLP (22.09.2015). We clustered this graph with respect to its modularity. Graph (a) shows the input graph and (b) shows the clustering with the highest modularity score. Graphs (c) and (d) visualize for members that have a joint publication, how likely they become merged into a same cluster when drawing samples from the corresponding distribution with different temperatures. Note that the positions and color of the nodes are for visualization only and do not represent features. Only the graph structure was taken into account for this particular problem.
Our approach identified 5 natural clusters among the program committee members. Each cluster is coherent in that its members either are located in a certain region of Europe and/or are members of a scientific subcommunity (scale space, mathematical imaging, computer vision). Furthermore, 6 members of the board were identified as being not connected to the community. For the reasons of anonymity, however, we do not go into further details. In the probabilistic setting, the approach suggests new edges. And each of these edges looks natural in that everyone who knows the community at large, would agree that the two persons that become connected are close to each other, for one or another reason. For example some relations between members of cluster 0 and 3 are more likely than others. This is caused by an reasonable cluster by some members of cluster 0 and 3, which does not show up in the optimal clustering. Relations between member of cluster 0 and 1 on the other hand, remain unlikely in the probabilistic setting. Such conclusions quite to the point are surprising in view of the fact that the approach only used the DBLP publication data as input.

While adding zero-weighted edges for the MAP-problem has no direct influence, we have shown that this is not the case for the probabilistic setting and the choice of the chosen graph-topology influences the space of partitions and in turn the probability of single edges to be between clusters. Consequently, the topology of a graph affects the stability of the MAP-solution.

For our problem settings we showed that the use of approximative solvers within the Perturb and MAP formulation leads to very similar results and the additional error is smaller that the one caused by using low-dimensional perturbation instead of global one. Since approximative methods often scale much better, this allows us to compute pseudo marginals very efficiently. As all samples can be calculated in parallel, the overall runtime of computing the marginals is only slightly larger than for the MAP-solution.

The availability of marginals can be used to guide the user in a interactive manner to regions with uncertain information. Furthermore, because the marginals are the derivation of the log partition function, the present work opens the door for probabilistic learning for clustering problems which takes probabilistic uncertainties into account.

**Acknowledgements** This work has been supported by the German Research Foundation (DFG) within the programme "Spatio-/Temporal Graphical Models and Applications in Image Analysis", grant GRK 1653.